\documentclass[letterpaper, 10 pt, conference]{ieeeconf}  
\IEEEoverridecommandlockouts                              
\usepackage{amsmath,amsfonts}
\usepackage{algorithm}
\usepackage{algpseudocode}
\usepackage{array}
\usepackage{textcomp}
\usepackage{stfloats}
\usepackage{url}
\usepackage{tabularx}
\usepackage{verbatim}
\usepackage{wrapfig}
\usepackage{adjustbox}
\usepackage{graphicx}
\usepackage{cite}
\usepackage{amsmath, amssymb}
\newcommand{\E}{\mathbb{E}}
\usepackage{tabulary,multirow,multicol,overpic}
\usepackage[table]{xcolor}
\usepackage{tabularx}
\usepackage[T1]{fontenc}
\usepackage{hyperref}
\usepackage{colortbl}
\usepackage{booktabs}

\DeclareMathOperator*{\argmin}{arg\,min}

\renewcommand{\arraystretch}{1.2}

\title{\LARGE \bf
\textbf{Efficient Continual Adaptation of Pretrained Robotic Policy with Online Meta-Learned Adapters}
}

\author{Ruiqi Zhu, Endong Sun, Guanhe Huang, Oya Celiktutan
\thanks{All authors are with the Department of Engineering, King's College London, London, UK (ruiqi.3.zhu@kcl.ac.uk).}
}

\begin{document}

\maketitle

\begin{abstract}

Continual adaptation is essential for general autonomous agents. For example, a household robot pretrained with a repertoire of skills must still adapt to unseen tasks specific to each household. Motivated by this, building upon parameter-efficient fine-tuning in language models, prior works have explored lightweight adapters to adapt pretrained policies, which can preserve learned features from the pretraining phase and demonstrate good adaptation performances. However, these approaches treat task learning separately, limiting knowledge transfer between tasks. In this paper, we propose Online Meta-Learned adapters (OMLA). Instead of applying adapters directly, OMLA can facilitate knowledge transfer from previously learned tasks to current learning tasks through a novel meta-learning objective. Extensive experiments in both simulated and real-world environments demonstrate that OMLA can lead to better adaptation performances compared to the baseline methods. The project link: \url{https://ricky-zhu.github.io/OMLA/}.

\end{abstract}


\section{Introduction}

In recent years, a series of pretrained models have emerged in the domains of computer vision, natural language processing, and speech recognition \cite{brown2020language, radford2021learning, radford2023robust, liu2024visual}. Pretrained on internet-scale data, these models exhibit strong zero-shot generalization and adapt effectively to downstream tasks with minimal task-specific data, which can be costly to collect \cite{zhu2023learning, zhu2024cross}. Similar practices have also been explored in robotic control applications \cite{brohan2022rt,brohan2023rt,team2024octo}. However, decision-making tasks in robotics present unique challenges due to the diverse and unstructured nature of real-world environments. Therefore, while these pretrained robotic policy models have shown the capability to extrapolate the behaviours to new initial conditions such as object positions or lightning, they struggle to solve tasks involving novel objects or unseen task instructions \cite{brohan2022rt,kim2024openvla}. 

To address this, robotic agents must retain the ability to adapt to new tasks that differ from earlier tasks. For example, pretrained household robots should be capable of \emph{continually adapting} to new tasks that involve novel instructions and previously unseen objects as required in each household. However, the question of how to efficiently and effectively continually adapt pretrained robotic policies to new tasks remains an open challenge. Conventional continual adaptation methods usually struggle with the balance of the issue of catastrophic forgetting and the issue of conservative learning \cite{kumar2022fine, wolczyk2021continual}.
To mitigate these issues in continual adaptations, inspired by parameter-efficient fine-tuning in language models, prior works have leveraged lightweight adapters (e.g. LoRA \cite{hu2021lora}, RoboAdapter \cite{sharma2023lossless}) to adapt pretrained models to new robotic tasks \cite{liu2023tail, kim2024openvla}. These adapters introduce a relatively small number of parameters compared to the whole model parameters while achieving effective adaptations on new tasks. Crucially, they mitigate catastrophic forgetting, as adapters are task-specific and operate in a plug-and-play fashion.


\begin{figure}[t!]
	\centering
	\includegraphics[width=\columnwidth]{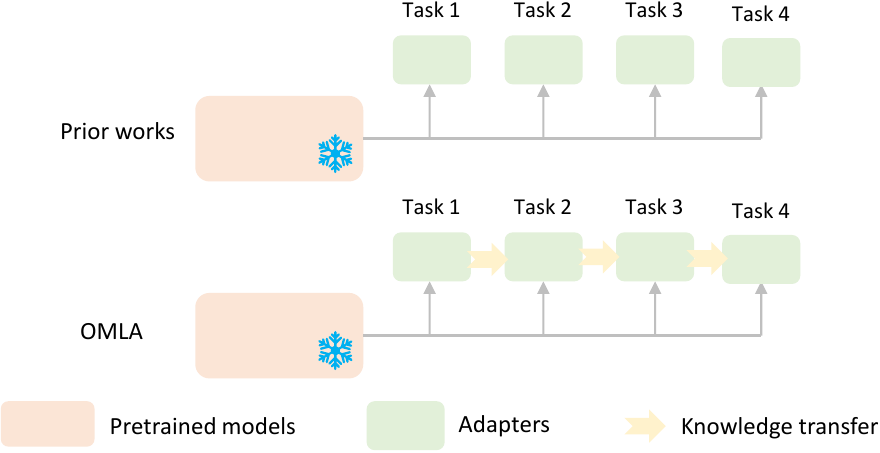}
	\caption{Our motivation: instead of learning adapters for each task separately, we develop OMLA for transfering the knowledge from previously learned tasks to facilitate new task adaptation.}
	\label{fig: motivation}
  \vspace{-0.5cm}
\end{figure}
 
In continual adaptations, beyond preventing catastrophic forgetting, knowledge transfer from previously learned tasks to current learning tasks is also essential \cite{diaz2018don,parisi2019continual,khetarpal2022towards}. However, these prior works treat the adapter learning across tasks separately as shown in Fig. \ref{fig: motivation}, limiting the knowledge transfer. This raises the question: \emph{How can we enable the knowledge transfer when using adapters to adapt pretrained robotic policies?}

In this paper, we propose \textbf{Online Meta-Learned Adapters (OMLA)} for the continual adaptation of pretrained robotic policies. Unlike prior works that train adapters for new tasks from scratch, OMLA can leverage data from previous tasks to learn a prior over adapter parameters, facilitating knowledge transfer and improving adaptation to new tasks. We evaluate our method extensively in both simulated and real-world environments, focusing on vision-language robotic policies. In simulated experiments, OMLA consistently outperforms baseline methods across diverse task setups and policy structures, even with limited task-specific demonstrations. In real-world experiments, policies trained with OMLA exhibit robust performances to unseen scenes, demonstrating its effectiveness in continual adaptation.

\section{Related Works}
\noindent\textbf{Continual Learning.}  
Continual learning in control is a long-studied yet challenging topic \cite{thrun1995lifelong, mccloskey1989catastrophic}. Its goal is to enable agents to acquire new tasks while maintaining the performance of previously learned ones. A key desideratum is leveraging prior knowledge or experience to accelerate new task learning \cite{parisi2019continual, khetarpal2022towards}. Most existing continual learning methods in robotic control require updates to the entire model parameters \cite{wolczyk2021continual, kaplanis2019policy, khetarpal2022towards}, making them prone to catastrophic forgetting or overly conservative updates that limit adaptability. In contrast, our method adopts task-specific lightweight adapters that maintain flexibility for new tasks while preventing forgetting previously learned tasks. In addition, rather than applying adapters in a straightforward manner \cite{liu2023tail}, we introduce a meta-learning objective that promotes effective knowledge transfer between tasks.

\noindent\textbf{Meta-learning in Robotics.} Meta-learning aims to discover useful priors from a collection of datasets to facilitate fast adaptation to new tasks \cite{finn2017model}. In robotics, meta-learning has been explored in contexts such as one-shot imitation learning \cite{finn2017one} and adaptation to domain shifts \cite{yu2018one, kaushik2020fast}. These approaches typically apply meta-learning objectives to entire robotic trajectories, which incurs high GPU memory costs—especially when dealing with high-dimensional observations (e.g., images) or large model architectures. We address this by proposing a more efficient strategy that applies meta-learning objectives selectively to informative data points by comparing the similarities of the data points, significantly reducing memory overhead, detailed in Section~\ref{sec:method:meta}.

\noindent\textbf{Adapters for pretrained models.} Adapters are a form of Parameter-Efficient Fine-Tuning (PEFT) originally developed for adapting large language models (LLMs) \cite{hu2021lora, houlsby2019parameter, li2021prefix}. These methods reduce computational overhead by fine-tuning a small set of additional parameters instead of modifying the entire model. Since adapters are task-specific, they naturally align with a key principle of continual learning: new task acquisition should not degrade performance on previously learned tasks. While prior works primarily employ adapters to mitigate catastrophic forgetting \cite{liu2023tail}, we extend their application by investigating how they can actively facilitate knowledge transfer across tasks, thereby enhancing continual learning performance.

\section{Background}
\noindent\textbf{Transformer.} Originally introduced for natural language processing by Vaswani et al. \cite{vaswani2017attention}, the Transformer architecture has since become the foundation of large models across various domains \cite{brown2020language, touvron2023llama, devlin2018bert}. The attention mechanism enables transformers to capture dependencies between tokens by assigning different weights across an input sequence, making them well-suited for processing long-context data. Transformer is further extended to tackle image data \cite{dosovitskiy2020image}. The Transformer architecture has also been extended to image data \cite{dosovitskiy2020image}. In this formulation, images are partitioned into patches, flattened, and encoded as tokens via linear projections. This enables the model to capture spatial relationships in images through attention maps. Moreover, Transformers can model temporal dependencies, making them well-suited for robotic policy learning, where they have achieved impressive performance \cite{chi2023diffusion, chen2021decision}.

\noindent\textbf{Meta-learning.} Meta-learning aims to bootstrap from a set of tasks to learn faster on a new task \cite{schmidhuber1987evolutionary}. To achieve that, Model-Agnostic Meta-Learning (MAML) \cite{finn2017model} learns a set of initial parameters. During meta-training, the model updates its parameters using a meta-training dataset, yielding a set of adapted parameters. Following that, gradients are computed using the adapted parameters and a meta-validation dataset. These gradients are propagated back to refine the initial parameters. Since this process involves differentiating through the adaptation step, it requires second-order derivatives. However, meta-learning generally assumes the tasks come from a fixed distribution, whereas continual learning operates in a non-stationary setting. To address this challenge, we build upon online meta-learning \cite{finn2019online}, which is detailed in Section \ref{sec:method:meta}.

\section{Method}

\subsection{Problem Formulation}
In the continual adaptation of agents, the agents pretrained on a set of tasks, need to be adapted to a sequence of new tasks continually. Beyond mitigating catastrophic forgetting of previously learned tasks, effective knowledge transfer is crucial for facilitating learning new tasks.
Each task $\mathcal{T}$ has an associated dataset in the form of $\{l^{\mathcal{T}},o_{t}^{\mathcal{T}},a_{t}^{\mathcal{T}}\}_{t=0}^{t=H}$, where $l^{\mathcal{T}}$ represents the task description, and $o_{t}^{\mathcal{T}}$ and $a_{t}^{\mathcal{T}}$ denote the observation and the action at time step $t$ respectively.

\subsection{Vision-language Robotic Policy Architecture} \label{sec:method:architecture}

\begin{figure}[t]
	\centering
    \vspace{-0.3cm}
	\includegraphics[width=\columnwidth]{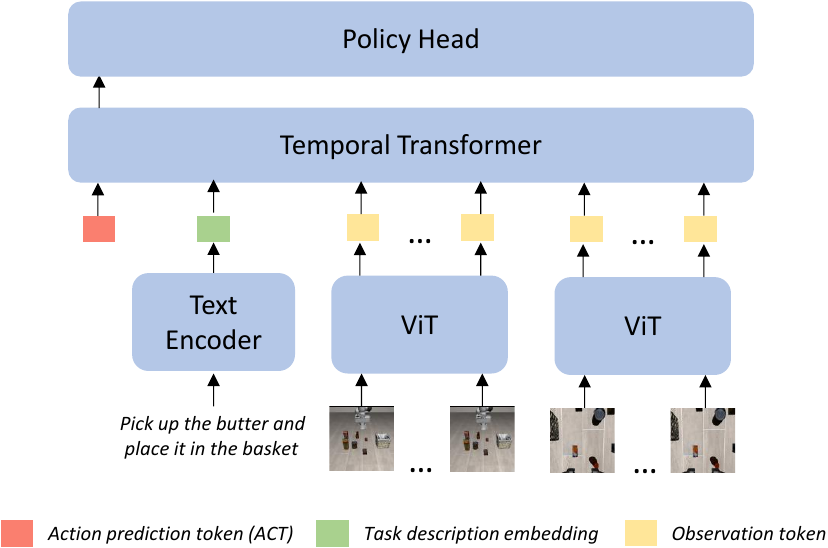}
	\caption{The vision-language policy architecture used in our experiments. The inputs consist of the task description, image observations and proprioceptive states. A history of observations spanning $C$ time steps is used. Following that, an action prediction token [ACT] is prepended to the token list before being processed by the temporal transformer. The output of the temporal transformer corresponding to the token [ACT] serves as the input to the policy head for action prediction.}
	\label{fig: architecture}
    \vspace{-0.3cm}
\end{figure}

Fig. \ref{fig: architecture} provides an overview of our vision-language robotic policy architecture. The model takes three types of inputs: a task description, visual observations, and the proprioceptive states of the robotic arm. The task description is encoded into task embedding with off-the-shelf text encoding models \cite{devlin2018bert, radford2021learning}. Proprioceptive states are mapped into the same embedding space as the task embedding using multilayer perceptrons (MLPs). A Vision Transformer (ViT), following prior works \cite{dosovitskiy2020image}, encodes the visual observations.

To infer the underlying states from observations, we construct a sequence of observations over $C$ consecutive time steps instead of relying on a single frame. The encoded features of the observations via ViT as a sequence of tokens are then fed into a temporal transformer module, which captures global dependencies across time steps to infer the underlying state.
Additionally, an action token [ACT] is prepended to the sequence of tokens. Thus, the input tokens to the temporal transformer module are then $\{[ACT], f^{l}, f^{\mathcal{O}}_{t},...,f^{\mathcal{O}}_{t+C-1} \}$, where $f^{\mathcal{O}}_{t}$ denotes the encoded observation at time step $t$ and $f^{l}$ represents the encoded task description. The output corresponding to the [ACT] token is subsequently passed to the policy head for the action prediction.

For the policy head, following prior works\cite{liu2024libero, liu2023tail}, we employ a Mixture of Gaussian Distributions to model the multi-modal action distributions. The final action is sampled from the predicted distribution. The loss function we use for the action prediction is the maximum likelihood estimation (MLE) loss defined below.
\begin{equation}
\label{eq:loss}
    \mathcal{L}(\theta) = \E_{(o_{t:t+C-1},a_{t:t+C-1})\sim \mathcal{D}}[-\sum_{i=t}^{i=t+C-1}\log \pi_{\theta}(a_{i}|o_{i},l)]
\end{equation}

\subsection{Adapting Pretrained Robotic Policy with Adapters}
In continual adaptation, it is essential to prevent catastrophic forgetting of previously learned tasks while leveraging prior knowledge to enhance learning in new tasks.
Conventional continual learning methods often struggle with balancing the issues of catastrophic forgetting and overly conservative learning \cite{wolczyk2021continual}. Inspired by parameter-efficient fine-tuning techniques in language
models, we employ adapters to adapt the pretrained robotic policies. Specifically, we adopt Low-Rank Adaptation (LoRA) \cite{hu2021lora} as adapters, since in prior works LoRA is reported to have better and more stable performance compared to other adapters in robotic policy adaptations \cite{liu2023tail}. LoRA is based on the insight that over-parameterized models often reside in a low-dimensional intrinsic subspace. LoRA adapts the pretrained model by introducing low-rank matrices $B$ and $A$ to modify the original weight matrix $W_{0}\in \mathcal{R}^{d\times k}$, as formulated in Eq.~\ref{eq:lora}. Here, $B\in \mathcal{R}^{d\times r}$ and $A\in \mathcal{R}^{r\times k}$ with $r\ll \min(d,k)$. Given an input $x\in \mathcal{R}^{d}$, the adapted output $\hat{h}$ is computed as follows:

\begin{equation}
\label{eq:lora}
    \hat{h} = W_{0}x + BAx. 
\end{equation}
 
Beyond parameter efficiency, a key advantage of using task-specific adapters in the continual adaptation of pretrained robotic policies is their ability to mitigate catastrophic forgetting of previously learned tasks. This property makes them more effective and flexible than conventional continual learning methods in learning new tasks.

\begin{algorithm}
\caption{Training with OMLA}
\label{algo:OMLA}
\begin{algorithmic}[1]
\State \textbf{Input:} Pretrained robotic policy models \texttt{$\pi_{\theta}(a_{t}|o_{t},l)$}, seen task dataset \texttt{$\mathcal{D}$}, task sequence $\mathcal{TS}$
\For{$\mathcal{T}$ in $\mathcal{TS}$}
    \State Initialize adapters and task-specific dataset $\mathcal{D}_{\mathcal{T}}$
    \For{$i = 1, 2, ...,$} \Comment{online meta-learning} 
        \State Update adapters $\phi_{\mathcal{T}}$ with $\mathcal{D}$ using Eq. \ref{eq:online meta learning}.
    \EndFor
    \For{$j = 1, 2, ...,$} \Comment{fine-tuning} 
        \State Fine-tuning online meta-learned adapters $\phi_{\mathcal{T}}$ 
        \State with $\mathcal{D}_{\mathcal{T}}$ using Eq. \ref{eq:loss}.
    \EndFor
\State \texttt{$\mathcal{D}$} $\gets$ \texttt{concat($\mathcal{D}$, $\mathcal{D}_{\mathcal{T}}$})
\EndFor

\end{algorithmic}

\end{algorithm}

\subsection{Online  Meta-Learned Adapters} \label{sec:method:meta}
Although adapters can be directly applied to fine-tune pretrained robotic policies, they do not inherently facilitate knowledge transfer from previously learned tasks. This limitation arises because adapters are initialized randomly and trained separately for each new task. To address this, we resort to meta-learning which has demonstrated the capability to extract the common structure from a collection of tasks, enabling fast adaptation to new tasks. However, meta-learning assumes access to a large collection of tasks, which is impractical in continual adaptation settings. To circumvent this limitation, we employ online meta-learning  \cite{finn2019online} to learn the adapters in the continual setting. The gradient computation of online meta-learning is shown in Eq. \ref{eq:online meta learning}. First, a task index $k$ is uniformly sampled from $h$ previously learned tasks $v^{h}$. The dataset corresponding to the task index is retrieved and constructed as meta-train dataset $\mathcal{D}_{k}^{tr}$ and meta-validation dataset $\mathcal{D}_{k}^{val}$. The adapter parameters denoted by $\phi$ are updated with gradients $g_{t}(\phi)$ computed from the loss function $\mathcal{L}$ shown as Eq. \ref{eq:loss} while keeping the pretrained model parameters frozen.

\begin{equation}
\label{eq:online meta learning}
\begin{aligned}
    &g_{t}(\phi)=\nabla_{\phi} \E_{k\sim v^{h}}\mathcal{L}(\mathcal{D}_{k}^{val},\mathcal{U}_{k}(\phi)), where\\
    &\mathcal{U}_{k}(\phi) \equiv \phi - \alpha\nabla_{\phi}\mathcal{L}(\mathcal{D}_{k}^{tr},\phi)
\end{aligned}
\end{equation}

\textbf{Proposed sampling strategy.} Prior works that utilize meta-learning for robotic policy learning usually meta-train on entire trajectories and meta-validate on separate ones \cite{finn2017one,yu2018one}. However, this approach is computationally expensive, particularly for image-based observations, and vision-language robotic policies that usually contain a large number of parameters, which can result in a large GPU memory requirement.
To mitigate these issues, we propose an alternative approach based on the insight that an agent should be able to predict an action for a given observation if it has encountered similar observations within a trajectory segment.

\begin{figure}[h!]
	\centering
	\includegraphics[width=\columnwidth]{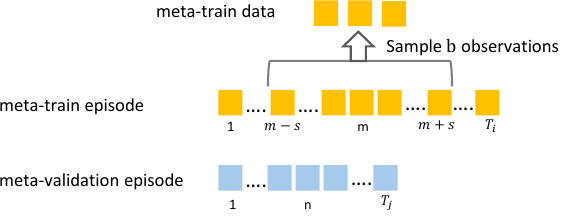}
	\caption{Illustration of the construction of meta-train data and meta-validation data. Instead of using one whole trajectory as the training dataset, which is computationally expensive, we sample the data points in the meta-train episode that are relevant to the data points in the meta-validation episode.}
	\label{fig: online}
    \vspace{-0.3cm}
\end{figure}

Specifically, we construct the meta-train data and meta-validation data based on the similarities between the datapoints. The solution is illustrated in Fig. \ref{fig: online}: Two episodes $i$ and $j$ of the selected task with the length of $T_{i}$ and $T_{j}$ are sampled from the dataset as the meta-train episode and meta-validation episode.
Then, an datapoint $(o_{n}^{j},a_{n}^{j})$ where the time step $n\sim Uniform[1,T_{j}]$ is sampled from the meta-validation episode as \emph{query} and is used as \textbf{meta-validation data}. Then we locate the the \emph{anchor} datapoint  $(o_{m}^{i},a_{m}^{i})$ in the meta-train episode by comparing the observation similarities using Eq. \ref{eq:anchor}, where $f_{ve}$ denotes the encoded visual feature. Notably, as the parameters of the pretrained robotic policies are frozen, observation features can be obtained before training therefore not affecting the learning process adversely.

\begin{equation}
\label{eq:anchor}
m =\argmin_{m'\in [1,T_{i}]}\lVert f_{ve}(o_{m'}^{i})-f_{ve}(o_{n}^{j}) \rVert_{2}
\end{equation}

After obtaining the \emph{anchor} in the meta-train episode, we randomly sample $b$ observations from the range $[m-s, m+s]$ as the \textbf{meta-train data}, where $s$ is the window size. In our experiment, the $b$ is set as $5$. Then we apply Eq. \ref{eq:online meta learning} with the meta-train data and the meta-validation data.
This strategy avoids the computational burden of processing entire episodes while preserving essential relevant information for online meta-learning.

Overall, instead of directly fine-tuning adapters from randomly initialized parameters, we construct a two-stage learning paradigm, as outlined in Algorithm \ref{algo:OMLA}. In the first stage, we utilize the online meta-learning objective to learn parameter priors, facilitating better initialization. In the second stage, the learned priors are fine-tuned using task-specific datasets, ensuring more efficient and effective adaptation.

\subsection{Implementation Details}
In the experiment, we use the BERT model \cite{devlin2018bert} to encode task descriptions. We use a two-layer MLP to encode proprioceptive states with a hidden size as 512. The policy head is modeled with a Gaussian Mixture Model with 5 modes, which is realized with a two-layer MLP with a hidden size as 1024. The observation horizon $C$ in our experiments is $10$. The rank of LoRA used in our experiments is 32. The vision-language policy model contains $16.5\times 10^6$ parameters while each adapter contains $0.23 \times 10^6$ parameters. The experiments are conducted on an NVIDIA 3090 graphic card with 24 GB GPU memory.

\section{Experiment}

We conduct experiments to answer the following questions: (Q1) Can our method OMLA draw benefits from previously learned tasks, leading to better adaptation performance? (Q2) How well can the performance of our approach scale with the dataset size? (Q3) Can our approach be robust across different policy architectures? (Q4) What are the impacts of the similarity between previously learned tasks and current learning tasks on adaptation performances?

\subsection{Experiment Setup}
\label{sec:exp_setup}
We have conducted quantitative evaluations in both physical simulation and real world. 

\begin{figure}[ht]
	\centering
    \vspace*{0.1cm}
	\includegraphics[width=\columnwidth]{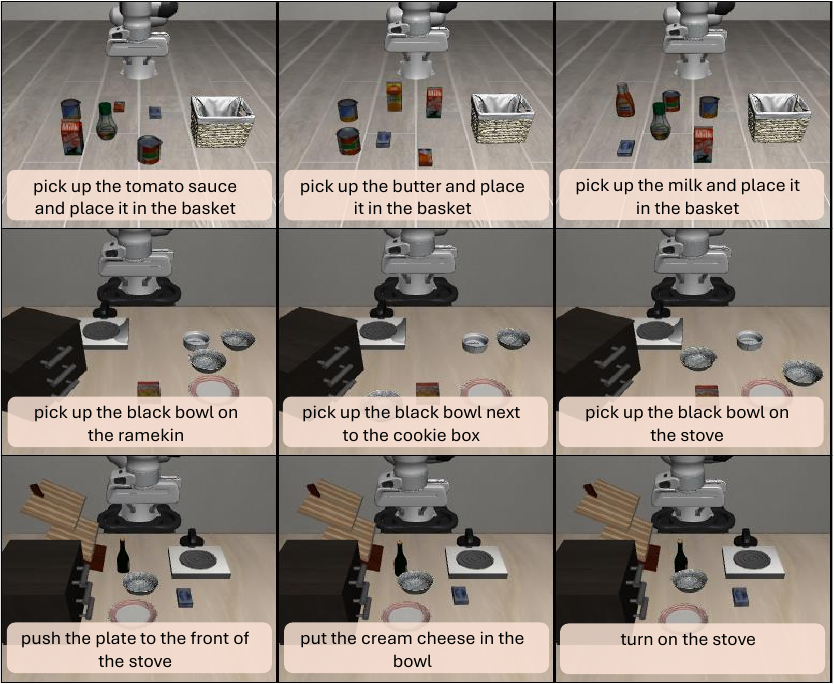}
	\caption{Examples of tasks of LIBERO (top to down: LIBERO-OBJECT, LIBERO-SPATIAL, LIBERO-GOAL).}
	\label{fig: task illustration}
  \vspace{-0.3cm}
\end{figure}

\noindent\textbf{Simulation.} For simulation experiments, we use a lifelong robot learning benchmark, LIBERO, as our testbed \cite{liu2024libero}. This benchmark is designed for robotic continual learning and provides diverse and challenging task setups. We use three task suites from the benchmark, LIBERO-OBJECT that contains pick-and-place tasks for different objects in the scene, LIBERO-SPATIAL that contains the same objects in each scene but with different spatial layouts, and LIBERO-GOAL that contains distinct goals (such as open the drawer, or turn on the stove). Each task suite consists of $10$ tasks. Examples of tasks from each of these suites are illustrated in Fig.~\ref{fig: task illustration}. The full task descriptions can be referred to \cite{liu2024libero}.             

The visual observations consist of RGB images captured from two viewpoints: the agent-view and the on-gripper view, both at the resolution of $128\times128\times3$. The proprioceptive state includes joint and gripper states. The action space consists of the translation, rotation, and gripper actions.

We evaluate our method on the three task suites, respectively. For each task suite, we use the first five tasks for pretraining the vision-language policies and then continually adapt the policy to the remaining five tasks. 

Each adaptation task is trained using 20 demonstrations. To assess robustness, we evaluate each adapted policy on 20 unseen initial scenes. Each evaluation is conducted with three different random seeds, and we report the average performance.

\begin{figure}[h!]

	\centering
	\includegraphics[width=\columnwidth]{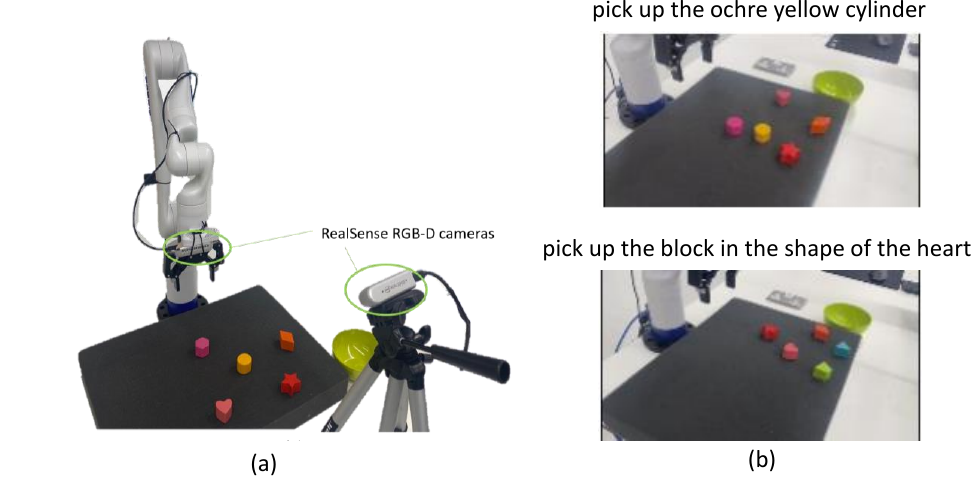}
	\caption{(a) The real robot experiment setup. (b) Real robot task examples.}
 \label{fig:real robot setup}
 \vspace{-0.3cm}
\end{figure}

\noindent\textbf{Real Robot Experiment.} For the real robot experiments, we focus on pick-up tasks with target objects of varying colors and shapes as shown in Fig. \ref{fig:real robot setup}. We use a Kinova robotic arm as the agent. Visual inputs are captured using two RealSense RGB-D cameras, providing agent-view and on-gripper-view images at a resolution of $640\times480\times3$. The proprioceptive state includes joint and gripper states. The action space includes translation, gripper control, and rotation, with rotation constrained to the yaw axis. We design 10 real-world tasks, each featuring one target object and four random distractor objects (Fig.~\ref{fig:real robot setup}a). The full task descriptions are presented in Table \ref{tab:real_task_suites}. The first five tasks are used for pretraining, while the remaining five are used for adaptation. Each adaptation task is trained with 20 demonstrations.

\begin{table}[t]
    \centering
    \caption{Task descriptions for simulated and real-world task suites}
    \renewcommand{\arraystretch}{1.2} 
    \setlength{\tabcolsep}{5pt} 
    \definecolor{lightgray}{gray}{0.9} 
    \begin{tabular}{|c|}
        \hline
        \rowcolor{lightgray}
        \textbf{Real Robot Tasks} \\
        \hline
        Pick up the blue pentagonal prism and place it in the bowl  \\
        Pick up the ochre yellow cross block and place it in the bowl  \\
        Pick up the orange rhombus prism and place it in the bowl \\
        Pick up the pink hexagonal prism and place it in the bowl  \\
         Pick up the red star block and place it in the bowl \\
          Pick up the ochre yellow cylinder and place it in the bowl\\
          Pick up the pink cube and place it in the bowl\\
          Pick up the pink heart block and place it in the bowl\\
          Pick up the green quarter cylinder and place it in the bowl\\
         Pick up the purple oblique rectangular prism and place it in the bowl\\
        \hline
    \end{tabular}
    \vspace{-0.5cm}
    \label{tab:real_task_suites}
\end{table}

\noindent\textbf{Evaluation Metrics.} Following prior work~\cite{liu2024libero}, we evaluate our method using two key metrics: \textbf{Forward Transfer} (FWT) and \textbf{Backward Transfer} (BWT). FWT refers to the highest performance when adapting to a new task. We denote FWT at task $k$ as $F_{k}$. BWT measures the success rate increase on previous tasks. Namely, when adapting to the $k-th$ task, we first record the best FWT model on this task and then evaluate this adapted model on all previous $k-1$ tasks, obtaining success rate $S_{i}$, $1\leq i \leq k-1$. Then we compute the success rate difference between the adapted model and the best FWT of the previous $k-1$ tasks and then average among them to obtain the BWT metric: $\mathbf{B} = \frac{1}{k-1} \sum_{i=1}^{k-1} (\mathbf{S}_i - \mathbf{F}_i)$. For both metrics, higher values indicate better performance.

\noindent\textbf{Baselines.} As in prior works, adapting pretrained robotic policies with LoRA is reported to have better performance compared to other adapters and conventional continual learning methods \cite{liu2023tail}, we mainly compare our method OMLA to \textbf{LoRA}. In addition, we also compare our method to \textbf{Experience Replay} (ER) \cite{chaudhry2019tiny} which employs a buffer for previously learned task datasets, using a 50-50 data split between new and previous learned tasks during new task training. Since LoRA and OMLA do not introduce impact on previously learned tasks, the BWTs of the both methods are zero.

\subsection{Experiment Results}

\begin{table*}[t]
\vspace{-0.3cm}
\centering
\caption{Continual Adaptation results on LIBERO (The BWT $\uparrow$ for LoRA and OMLA are all 0 (no catastrophic forgetting)).}
\label{table:libero-results}
  \begin{tabular}{l|cc|c|c|cc|c|c|cc|c|c}
    \hline
    \multirow{3}{*}{Tasks} & \multicolumn{4}{c|}{LIBERO-OBJECT} & \multicolumn{4}{c|}{LIBERO-SPATIAL} & \multicolumn{4}{c}{LIBERO-GOAL} \\
     & 
      \multicolumn{2}{c}{ER} & 
      \multicolumn{1}{c}{OMLA} &
      \multicolumn{1}{c|}{LoRA} & 
      \multicolumn{2}{c}{ER} & 
      \multicolumn{1}{c}{OMLA} &
      \multicolumn{1}{c|}{LoRA} & 
      \multicolumn{2}{c}{ER} & 
      \multicolumn{1}{c}{OMLA} &
      \multicolumn{1}{c}{LoRA}\\ 
      & FWT $\uparrow$ & BWT$\uparrow$ & FWT $\uparrow$ & FWT $\uparrow$& FWT $\uparrow$ & BWT$\uparrow$ & FWT $\uparrow$ & FWT $\uparrow$& FWT $\uparrow$ & BWT$\uparrow$ & FWT $\uparrow$ & FWT $\uparrow$\\
      \hline
      Task 6 & 0.55 & - & 0.74 & 0.65 & 0.52 & -& 0.68 & 0.48& 0.32& -& 0.28 & 0.20\\
      \hline
      Task 7 & 0.65 & -0.25 & 0.83 & 0.78& 0.58&-0.23 & 0.77 & 0.73&0.26 & -0.14& 0.42 & 0.30\\
      \hline
      Task 8 & 0.58& -0.40 & 0.91 & 0.66&0.44 &-0.37 & 0.70 & 0.58& 0.55& -0.18& 0.87 & 0.93\\
      \hline
      Task 9 & 0.61& -0.51& 0.88 & 0.68&0.70 &-0.44 & 0.78 & 0.66& 0.67& -0.25& 0.88 & 0.85\\
      \hline
      Task 10 & 0.55& -0.58& 0.93 & 0.75& 0.26& -0.48& 0.51 & 0.41&0.34 & -0.44& 0.45 & 0.32\\
      \hline
      Average &0.59&-0.43&\textbf{0.86}&0.71&0.37&-0.38&\textbf{0.69}&0.57 &0.43&-0.25&\textbf{0.58}&0.52\\
    \hline
  \end{tabular}
\end{table*}

\subsubsection{Simulation results}

The simulation results for the three task suites are summarized in Table \ref{table:libero-results}. Our method, OMLA, consistently outperforms all baselines across these suites. Experience Replay (ER) shows negative BWT, indicating forgetting of previously learned tasks. In addition, ER presents degraded FWT compared to LoRA and OMLA. This suggests that merely incorporating data from prior tasks does not effectively leverage the useful information within these datasets and can hinder the learning of new tasks.

\begin{figure}[h]
	\centering
	\includegraphics[width=\columnwidth]{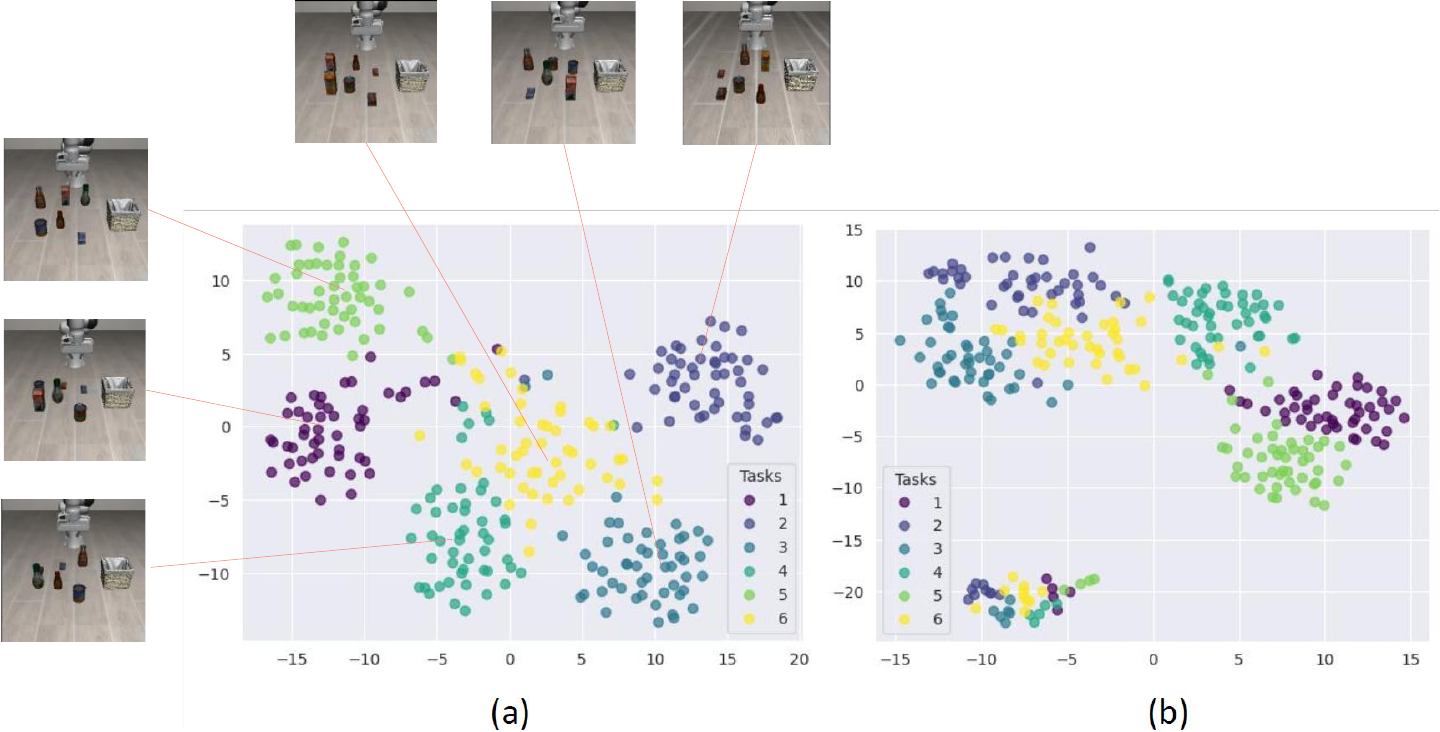}
	\caption{The representation of observations of different tasks. (a) after online meta-learning phase, (b) before online meta-learning pahse.}
	\label{fig: params}
    \vspace{-0.2cm}
\end{figure}

To further understand why OMLA surpasses LoRA, we analyzed the learned representations by randomly sampling unseen initial observations from both the pretraining and adapted task datasets. We visualized these representations using t-distributed stochastic neighbor embedding (t-SNE) before and after the online meta-learning phase, as shown in Fig. \ref{fig: params}. Prior to online meta-learning, the representations are highly entangled. Post the online meta-learning, they become more structured and well-separated, facilitating the subsequent fine-tuning phase. This observation aligns with findings in prior works which demonstrate that effective clustering of task embeddings correlates with improved adaptation performances \cite{li2024efficient}.

\subsubsection{Impact of dataset size}
To evaluate the effect of dataset size on performance, we vary the demonstration numbers to $5$ and $10$. The average performances across task suites for both OMLA and LoRA are reported in Table \ref{table:demo variations}. 
We observe that our method OMLA outperforms LoRA by approximately $\textbf{20\%}$ consistently with different numbers of demonstrations. 

\begin{table}[h]
\centering
\caption{Continual adaptation with varing demonstration numbers}
\label{table:demo variations}
  \begin{tabular}{l|ccc|ccc}
  \hline
    \multirow{2}{*}{Task Suites} & \multicolumn{3}{c}{OMLA} & \multicolumn{3}{c}{LoRA} \\
    & 5&10&20& 5&10&20\\
    \hline
    LIBERO-OBJECT&0.58&0.73&0.86&0.48&0.61&0.71\\
    LIBERO-SPATIAL&0.46&0.60&0.69&0.35&0.48&0.57 \\
    LIBERO-GOAL&0.27&0.41&0.58&0.24&0.34&0.52 \\
    \hline
    Average& \textbf{0.44}&\textbf{0.58}&\textbf{0.71}&0.36&0.47&0.60\\
    \hline
  \end{tabular}
  \vspace{-0.3cm}
\end{table}

\subsubsection{Impact of policy architecture}
To assess the effectiveness of OMLA across different policy architectures, we evaluate both OMLA and LoRA using an alternative architecture, denoted as \textbf{A1}, while the primary architecture is referred to as \textbf{A0}. The key difference between A0 and A1 lies in the visual encoder: A0 employs a Vision Transformer (ViT), whereas A1 utilizes a ResNet-18 backbone \cite{he2016deep}. The results are presented in Table \ref{table:policy arc variations}.

Compared to A0, the reduced expressivity of A1 leads to a decline in overall performance. However, OMLA continues to outperform LoRA across both architectures, demonstrating its robustness to variations in policy design.

\begin{table}[h]
\centering
\caption{Continual adaptation with Varied policy architectures \textbf{A1}}
\label{table:policy arc variations}
  \begin{tabular}{l|ccc|ccc}
  \hline
    \multirow{2}{*}{Task Suites} & \multicolumn{3}{c}{OMLA} & \multicolumn{3}{c}{LoRA} \\
    & 5&10&20& 5&10&20\\
    \hline
    LIBERO-OBJECT&0.48&0.66&0.75&0.38&0.51&0.63\\
    LIBERO-SPATIAL&0.19&0.30&0.43&0.12&0.21&0.38 \\
    LIBERO-GOAL&0.16&0.22&0.34&0.07&0.19&0.23 \\
    \hline
    Average& \textbf{0.28}&\textbf{0.39}&\textbf{0.51}&0.19&0.30&0.41\\
    \hline
  \end{tabular}
\end{table}

\subsubsection{Impact of previously learned tasks} 
To assess how previously learned tasks influence current task learning, we have designed two scenarios and tested them on the LIBERO-OBJECR task suite. The two scenarios are: (S1) to assess the impact of similarity between previously learned tasks and current learning tasks, we modified the pretraining task dataset by replacing it with the dataset of the first five tasks from LIBERO-GOAL. This adjustment introduces a discrepancy between the previously learned tasks and the current learning tasks, (S2) to evaluate the importance of incorporating task datasets during the continual adaptation phase, we exclude these datasets and retain only the pretraining task datasets for online meta-learning. We report the results in Table \ref{table:variation}. For the ease of comparison, we include the results of OMLA and LoRA under the original setting as discussed in Section \ref{sec:exp_setup}, denoted as OMLA and LoRA in the table respectively. We observe that: (1) performance in S1 decreases by $14.6\%$ compared to OMLA, indicating that the similarity between previously learned tasks and current learning tasks is critical for effective knowledge transfer, (2) the performance in S2 decreases by $7.5\%$ compared to OMLA, highlighting the importance of dataset incorporation during the continual adaptation phase, (3) though the performance drops compared to OMLA under the original setting, the performances under S1 and S2 still outperform the performances of LoRA.

\begin{table}[h]
\vspace{-0.2cm}
\centering
\caption{Continual adaptation on libero-object with 2 different scenarios}
\label{table:variation}

  \begin{tabular}{l|>{\columncolor[gray]{0.9}}c>{\columncolor[gray]{0.9}}ccc}
  \hline
    Scenarios &OMLA& LoRA&S1 & S2 \\
    \hline
    Task 6 & 0.74&0.65&0.67 & 0.74 \\
    \hline
    Task 7 & 0.83&0.78&0.75 & 0.80 \\
    \hline
    Task 8 & 0.91&0.66&0.77 & 0.82 \\
    \hline
    Task 9 & 0.88&0.68&0.82 & 0.85 \\
    \hline
    Task 10 & 0.93&0.75&0.76 & 0.81 \\
    \hline
    Average& \textbf{0.86}&0.71&0.75&0.80\\

    \hline
  \end{tabular}

\vspace{-0.2cm}
\end{table}

\subsubsection{Real robot experiments}
In this section, we evaluate the performance of OMLA and LoRA in real-world environments under varying demonstration sizes (10 and 20). Each adapted policy is tested across 10 rollouts, each featuring an unseen layout. During the evaluation, we observe that the policies adapted with LoRA fail to recognize the target objects and exhibit jittery motions in certain unseen scenes. In contrast, policies adapted with OMLA generally perform the tasks smoothly. The quantitative results of real robot experiments are summarized in Table \ref{table:real-robot}. These results demonstrate that OMLA outperforms LoRA in real robot experiments with varing demonstration numbers consistently.

\begin{table}[ht]
\centering
\caption{Continual single-task adaptation performance of real robot experiments.}
\label{table:real-robot}
  \begin{tabular}{l|cc|cc}
    \hline
    \multirow{2}{*}{Tasks} 
    & \multicolumn{2}{c}{OMLA} & \multicolumn{2}{c}{LoRA}   \\ 
    & 10& 20  & 10& 20  \\
    \hline
    Task 6 & 6/10 & 9/10 & 3/10 & 6/10   \\ 
    \hline
    Task 7  & 5/10 & 8/10 & 7/10 & 8/10   \\
    \hline
    Task 8  & 5/10 & 8/10 & 3/10 & 5/10  \\
    \hline
    Task 9  & 3/10 & 9/10 & 1/10 & 4/10   \\
    \hline
    Task 10  & 5/10 & 8/10 & 3/10 & 7/10   \\
    \hline
    Average  & \textbf{48.0 \%} & \textbf{84.0} \% & 34.0 \% & 60.0 \%   \\
    \hline
  \end{tabular}
  \vspace{-0.3cm}
\end{table}
\section{Conclusion}
In this paper, we presented OMLA, an approach for the efficient continual adaptation of pretrained robotic policies. OMLA adopts lightweight adapters to adapt pretrained robotic, preventing catastrophic forgetting issues since the adapters are task-specific. In addition, we have introduced an online meta-learning objective for discovering adapter parameter priors with the datasets of previously learned tasks to facilitate new task learning.
Experiments in both simulated and real-world environments demonstrate that OMLA outperforms baseline models. While our proposed approach demonstrates promising results in continually adapting pretrained robotic policies with adapters, several potential extensions remain outside the scope of this work. First, the way to identify anchors in meta-train episodes could be improved by incorporating context understanding of the task descriptions. Another direction for improvement is to enhance the policy architecture by incorporating the diffusion policy head, which excels at multi-modal distribution predictions. We leave these directions for future work.

\bibliographystyle{IEEEtran}
\bibliography{main}

\end{document}